\documentclass[runningheads]{llncs}
\usepackage{enumitem}
\usepackage{multirow}
\usepackage{makecell}
\usepackage{amsfonts}
\usepackage{bm}
\usepackage{thmtools}  
\usepackage{amsmath}

\usepackage{graphicx}
\usepackage{enumitem}
\usepackage{multirow}
\usepackage{makecell}
\usepackage{hyperref}
\usepackage{amsfonts}
\usepackage{bm}
\usepackage{xcolor}
\usepackage{thmtools}  
\usepackage{booktabs}
\usepackage{amsmath}
\usepackage[ruled,vlined]{algorithm2e}
\usepackage{dsfont}
\usepackage{tabularx}
\usepackage{makecell}

\usepackage{graphicx}
\hypersetup{
colorlinks=true,
linkcolor=black
}
\usepackage{graphicx}
\newcommand{\ours}[0]{\textsc{ORAL}}

\newcommand{\ourtopic}[0]{Open-world Graph Learning}
\newcommand{\revise}[1]{\textcolor{black}{#1}}

\hypersetup{
colorlinks=True,
linkcolor=blue,
urlcolor=blue,
citecolor=blue
}
\begin{document}
\title{Beyond the Known: Novel Class Discovery for Open-world Graph Learning}
%
%
\author{Yucheng Jin\inst{1} \and
Yun Xiong \inst{1} \and Juncheng Fang \inst{1} \and Xixi Wu \inst{1} \and Dongxiao He \inst{2} \and Jia Xing \inst{1} \and Bingchen Zhao \inst{3} \and Philip Yu \inst{4}}
\institute{Shanghai Key Laboratory of Data Science, School of Computer Science, Fudan University, China \\ \email{\{22110240021,23212010010,21210240043\}@m.fudan.edu.cn, \{yunx,18110240037\}fudan.edu.cn} \and College of Intelligence and Computing, Tianjin University \\ \email{hedongxiao@tju.edu.cn} \and School of Informatics, University of Edinburgh  \\ \email{zhaobc.gm@gmail.com} \and Department of Computer Science, University of Illinois at Chicago 
\email{psyu@cs.uic.edu}}
\maketitle              
\begin{abstract}
Node classification on graphs is of great importance in many applications. Due to the limited labeling capability and evolution in real-world open scenarios, novel classes can emerge on unlabeled testing nodes. However, little attention has been paid to novel class discovery on graphs. \revise{Discovering novel classes is challenging as novel and known class nodes are correlated by edges, which makes their representations indistinguishable when applying message passing GNNs. Furthermore, the novel classes lack labeling information to guide the learning process. In this paper, we propose a novel method \underline{O}pen-world g\underline{R}\underline{A}ph neura\underline{L} network (\ours{}) to tackle these challenges. \ours{} first detects correlations between classes through semi-supervised prototypical learning. Inter-class correlations are subsequently eliminated by the prototypical attention network, leading to distinctive representations for different classes. Furthermore, to fully explore multi-scale graph features for alleviating label deficiencies, \ours{} generates pseudo-labels by aligning and ensembling label estimations from multiple stacked prototypical attention networks.} Extensive experiments on several benchmark datasets show the effectiveness of our proposed method.

\end{abstract}
\section{Introduction}
Graphs are ubiquitously used to reveal the interactions among various entities. One fundamental learning task on graphs is the node classification \cite{huang2018adaptive,Izadi2020OptimizationOG,kipf2016semi,luan2021heterophily,biology}, which has many applications such as predicting the areas of publications in academic networks \cite{Izadi2020OptimizationOG,luan2021heterophily}, identifying the functions of proteins in biology graphs \cite{Hetzel2021GraphRL,biology}. Graph neural networks \cite{Bruna2013SpectralNA,Niepert2016LearningCN}, which rely on extensive annotation information for training, have shown superior performance on the node classification task. To further ease the annotation burden, researchers have recently paid much attention to the semi-supervised setting \cite{Chen2020ConvolutionalKN,Gao2019ExploringSG,kipf2016semi,Liao2018GraphPN}, where only a small proportion of nodes on the graph is labeled.  

\begin{figure}[t]
    \centering
    \includegraphics[width=0.8\linewidth]{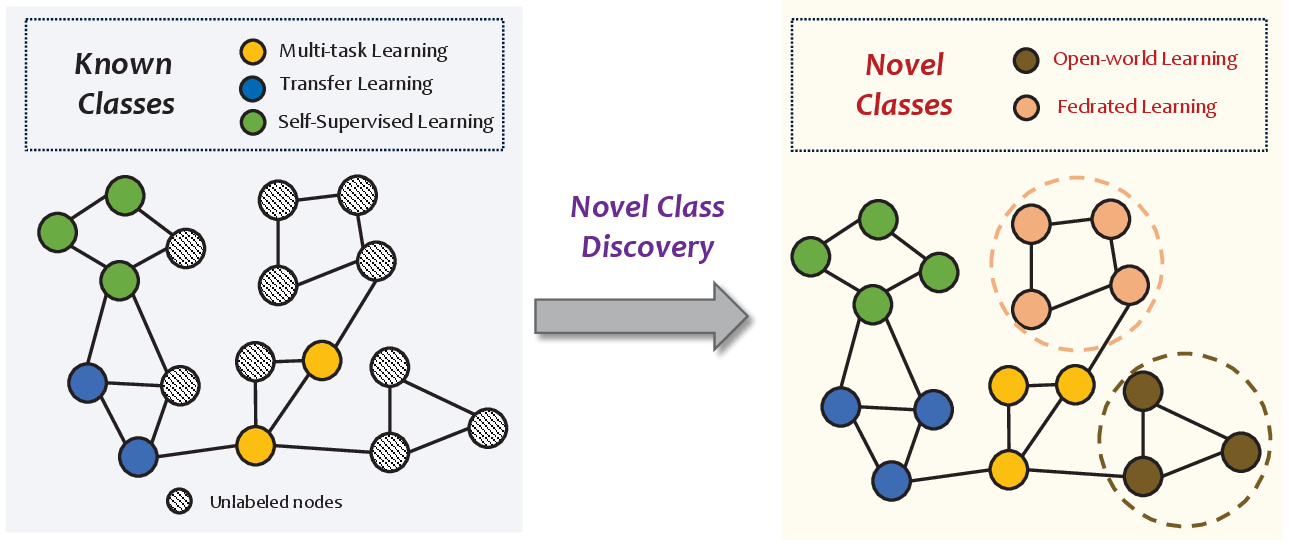}
    \caption{Illustration of novel class discovery for open-world learning on an academic graph. Nodes represent papers, edges represent citation relationships, and each paper belongs to a certain research field (node class).}
    \label{Figure open-world}
\end{figure}

However, existing node classification methods \cite{Bruna2013SpectralNA,huang2018adaptive,Izadi2020OptimizationOG,kipf2016semi,luan2021heterophily,biology,Niepert2016LearningCN} assume a closed-world setting \cite{Cao2021OpenWorldSL}, where unlabeled testing nodes belong to classes seen on the labeled training nodes. This assumption rarely holds for graphs in the open world, since novel classes can frequently emerge on unlabeled nodes. Take the toy academic graph in Figure \ref{Figure open-world} as an example, with nodes and edges denoting papers and citation relationships respectively. In addition to nodes labeled with well-known research topics, there are newly emerging unlabeled papers with unknown themes. The emergence of new research papers can introduce novel research topics. \revise{A crucial problem is to learn distinguishable representations for both known and novel classes, and facilitates novel class discovery along the way.} We term this problem as \textbf{\ourtopic{}} in this paper.

While the open-world graph learning naturally suits many real-world applications, it poses unique challenges: (1) Novel and known classes are correlated by edges, which makes learning distinguishable representations difficult. For example, papers proposing new research areas will cite existing articles in related fields on the academic graph. Consequently, novel and existing research areas are correlated by edges representing citation relationships. Edges between novel and known classes can make their representations indistinguishable, since typical GNNs \cite{Bruna2013SpectralNA,huang2018adaptive,Izadi2020OptimizationOG,kipf2016semi,Niepert2016LearningCN} tend to learn similar representations for connected nodes through message passing. (2) Only unlabeled nodes of novel classes are provided, lacking labeling information to guide representation learning and novel class discovery during the training process. Generating pseudo-labeling on novel class nodes is challenging due to the absence of labeled novel class nodes for learning the labeling mechanism. Given known class labeling information and rich features on graphs, how can we obtain novel class pseudo-labels and incorporate the generated label information for open-world graph learning? 





In this paper, we propose a novel method \underline{O}pen-world g\underline{RA}ph neura\underline{L} network (ORAL) to tackle these challenges. \ours{} first employs the prototypical attention network to detect and eliminate correlations between novel and known classes. By leveraging labeling granularity mined from known classes, the prototypical attention network first approximates the ground-truth label with semi-supervised prototypical clustering. Distinctive representations for different classes can be obtained by eliminating message passing over inter-class edges detected from label estimations. \ours{} stacks multiple prototypical attention networks to explore graph features at different scales. To alleviate the lack of novel class labeling information,  label estimations from different layers are aligned and ensembled to generate reliable pseudo-labels. Pseudo-labeling information is incorporated into the input graph data to guide the novel class discovery by recovering intra-class edges and removing inter-class ones. Confidence filtering and perturbation-robust consistency training are further introduced to combat pseudo-labeling noise during the learning process.


In summary, our contributions can be summarized as,
\begin{itemize}
    \item Our approach constitutes the first attempt to investigate open-world graph neural network which can automatically discover novel classes on graphs.
    \item We propose the prototypical attention network to detect and eliminate inter-class correlations for distinguishing novel classes from others during representation learning.
    \item \revise{To ease the lack of novel class label information, we generate pseudo-labels according to multi-scale graph features and labeling granularity mined from known classes.}
    \item We perform extensive experiments and analyze the results, proving the effectiveness of our method.
\end{itemize}

\begin{figure}[t]
    \centering
    \includegraphics[width=\linewidth]{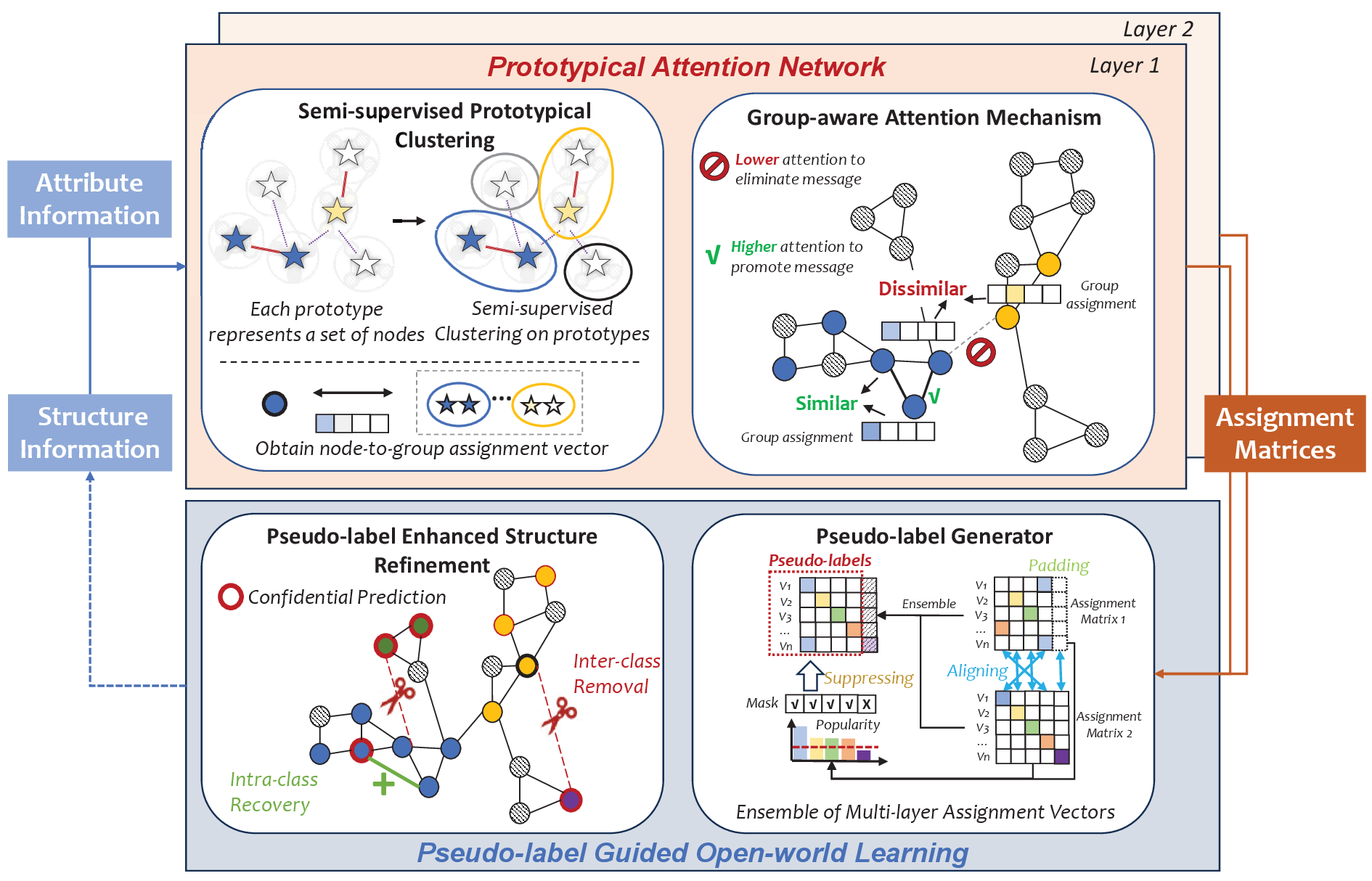}
    \caption{An overview of the proposed model \ours{}.}
    \label{Figure model}
\end{figure}

\section{Problem Formulation}
\textbf{Notations.} A graph is represented as $G=(\mathcal{V},\mathcal{E},\mathbf{X})$, where $\mathcal{V}=\{v_i\}_{i=1,...,N}$ is a node set representing nodes in a graph, $e_{ij}=(v_i,v_j) \in \mathcal{E}$ is an edge indicating the relationship between two nodes and $\mathbf{x}_i \in \mathbf{X}$ indicates content features associated with each node $v_i$. 

\subsubsection{\ourtopic{}.} The node set $\mathcal{V}$ is composed of labeled training nodes $\mathcal{V}_{L}$ and unlabeled testing nodes $\mathcal{V}_{U}$, which belong to class set $\mathcal{Y}^L$ and $\mathcal{Y}^U$ respectively. In open-world scenarios, novel class can emerge on the unlabeled nodes, i.e., $\mathcal{Y}^L \subset \mathcal{Y}^U$. The aim of open-world graph learning is to discover novel classes automatically and classify the unlabeled nodes into known classes ($\mathcal{Y}^L$) or novel classes ($\mathcal{Y}^U \setminus \mathcal{Y}^L$).

\section{Methodology}

\noindent \textbf{Overview.} The framework of our proposed method \ours{} is shown in Figure \ref{Figure model}. \revise{The prototypical attention network (PAN) is first employed to eliminate inter-class correlations. Given representative nodes (prototypes) extracted by prototype learning, PAN performs semi-supervised clustering on the prototype graph of relatively small size, efficiently obtaining group assignments of prototypes and nodes. Message passing between groups are eliminated by the group-aware attention mechanism to generate separable representations.} \revise{Furthermore, \ours{} eases the lack of labeling information by the pseudo-label generator, which aligns and ensembles multiple group assignment estimations in the last stage, facilitating the open-world learning through structure refining.}

\subsection{Prototypical Attention Network}
\revise{The prototypical attention network clusters nodes into groups and eliminates correlations between  classes by reducing message passing between groups, where the clustering results are approximations of ground-truth labels. Nonetheless, the number of novel classes is unknown in the open world, we have to repeatedly perform clustering on graphs to search for the optimal clustering granularity by fitting the labeling information, which can be time-consuming. To ease the computation burden, we extract representative nodes through prototype learning and conduct semi-supervised clustering on the small prototype graph instead of the large whole graph. The node clustering result is approximated from that of prototypes, which is employed to eliminate message passing through the group-aware attention mechanism.}

Representative nodes are modeled as a set of trainable prototypes $\mathcal{C} = \{\mathbf{c}_1, \mathbf{c}_2, \cdots \mathbf{c}_{N_{pro}} \}$ in the representation space, where $N_{pro}$ denotes the number of prototypes. Given a node with representation $\mathbf{h}_i$ which is initialized as node feature $\mathbf{x}_i$, we compute the representativeness score of each prototype $\mathbf{c}_j$ to the node as follows,
\begin{equation}
    r_{ij} = \frac{\exp({\mathbf{h}_i^T \mathbf{c}_j})}{\sum_{k=1}^{N_{pro}}{\exp({\mathbf{h}_i^T \mathbf{c}_k}})}.
    \label{Eq n2p}
\end{equation}
To ensure each prototype is representative of a set of nodes, we regularize the marginal distribution of representativeness scores to be close to a balanced prior,
\begin{equation}
    \mathcal{L}_{reg}=\mathrm{KL}[(\frac{1}{N_{pro}} \cdot \mathbf{1}) || (\frac{1}{|\mathcal{V}|}\sum_{i=1}^{|\mathcal{V}|}\mathbf{r}_i)],
    \label{loss reg}
\end{equation}
where $\mathbf{r}_i=[r_{i1},\ldots,r_{iN_{pro}}]\in \mathbb{R}^{N_{pro}}$, $\mathbf{1} \in \mathbb{R}^{N_{pro}}$ represents the all-ones vector and $\mathrm{KL}$ represents the Kullback-Leibler divergence. Each node is regarded as an associated instance of prototypes with the top $k$ largest representativeness score. The similarity relationships between prototypes $\mathbf{s}\in \mathbb{R}^{N_{pro}\times N_{pro}}$ can be calculated as the jaccard index of their associated node set, i.e., $s_{ij}=\text{Jaccard}(\Gamma_i,\Gamma_j)$, with $\Gamma_i$ denotes the associated node set of the $i$-th prototype. 

\subsubsection{Semi-supervised Prototypical Clustering.} Applying a clustering algorithm with polynomial time complexity on the small prototype graph $(\mathcal{C},\mathcal{E}^p)$ instead of the whole graph can substantially ease the computational burden, where $\mathcal{E}^p=\{(\mathbf{c}_i,\mathbf{c}_j,{s}_{ij})\}$. Given the groups $\mathcal{C}^g=\{\mathcal{C}_1^g,\cdots,\mathcal{C}_{N_{clu}}^g\}$ discovered by clustering on prototypes, the assignment probability of a node $v_i$ to a group $\mathcal{C}_k^g$ can be approximated according to their relationship scores with prototypes,
\begin{equation}
    p_{ik} = \sum_{\mathbf{c}_j \in \mathcal{C}^g_k}r_{ij}.
    \label{Eq groupassign}
\end{equation}
Despite prototype-based clustering can improve efficiency significantly, the labeling granularity which decides the class (cluster) number is still unknown. However, overly fine-grained clustering can cluster nodes of the same class into different clusters, while overly coarse-grained clustering can cluster nodes of different classes into the same cluster. Both cases lead to drops in clustering accuracy. 
Therefore, we search for the optimal clustering granularity by maximizing the accuracy on the labeled known class nodes. The clustering accuracy is calculated as, 
\begin{equation}
\begin{aligned}
\mathrm{ACC} = \mathop{\max}_{f\in \mathcal{P}(\mathcal{Y}^L)}&\sum_{v_i \in \mathcal{V}_L}\frac{1}{|\mathcal{V}_L|}\mathds{1}(\hat{y}_i=f(y_i)), \\
\hat{y}_i&=\mathop{\arg\max}\limits_{k}p_{ik},
\label{Eq acc}
\end{aligned}
\end{equation}
where $\mathds{1}$ is the indicator function and $\hat{y}_i$ is the predicted group ID. Considering the potential misalignment between between class and groups IDs, we find the alignment function from the set of all permutations $\mathcal{P}(\mathcal{Y}^L)$ by solving the maximization problem through the Hungarian algorithm \cite{Kuhn1955TheHM}. Groups do not match with any known class are regarded as discovered novel classes. 

\subsubsection{Group-aware Attention Mechanism.} 
Besides local attribute information on nodes, the neighborhood information on graphs has also been shown to be important for node classification \cite{Bruna2013SpectralNA,huang2018adaptive,Izadi2020OptimizationOG,kipf2016semi,Niepert2016LearningCN}. However, due to the existence of edges between known class nodes and novel class ones, aggregating all neighborhood information with existing graph convolution layers can make different classes indistinguishable in the representation space. Despite graph attention networks can assign different weights to different neighbors, which are not guaranteed to contain knowledge about how to distinguish each novel class from others.

To learn compact representations for novel class discovery, we propose the group-aware attention mechanism. Group assignments predicted by semi-supervised prototypical learning are approximations of the ground-truth labels, which can be utilized to eliminate correlations between novel and known classes during neighborhood aggregation. The group-aware attention mechanism calculates the attention scores for neighbors according to the group assignment similarities $e_{ik} = \cos{(\mathbf{p}_i, \mathbf{p}_k)}$, where $\mathbf{p}_i=[p_{i1},\ldots,p_{iN_{clu}}]$,
\begin{equation}
    \begin{aligned}
        \alpha_{ij} = \mathrm{softmax} & (e_{ij})=\frac{\exp{(e_{ij})}}{\sum_{k\in N(i)} \exp{(e_{ik})}}. \\
    \end{aligned}
\end{equation}
The low group assignment similarities between novel and known class nodes lead to low attention weights, while the attention weights between nodes from the same class are high. Consequently, aggregating information according to the group assignments can eliminate correlations between classes due to the low attention weights on inter-class edges, obtaining compact representations,
\begin{equation}
    \mathbf{h}_i^* = \sigma(\sum_{v_j\in N(i)\cup\{v_i\}} \alpha_{ij} \mathbf{W}\mathbf{h}_j).
    \label{Eq groupatt}
\end{equation}
where $\sigma$ is the relu activation function. By stacking multiple prototypical attention networks, \ours{} further predicts over the representation $\mathbf{h}_i^*$ obtained after selectively aggregating neighborhood information. Consequently, we get access to a list of predictions as well as node representations, $\{(\mathbf{p}^{(1)},\mathbf{h}^{(1)}),...,(\mathbf{p}^{(L)},\mathbf{h}^{(L)})\}$, calculated at different scales using different hops of neighborhood information, where $L$ is the number of layers, $(\mathbf{p}^{(i)},\mathbf{h}^{(i)})$ is the group assignments and representations generated by the $i$-th prototypical attention network. The prototypical attention networks are trained by the cross-entropy loss layer by layer,
\begin{equation}
    \mathcal{L}_{ce} = \sum_{v_i\in\mathcal{V}}\sum_{l=1}^L-\log p^{(l)}_{iy_i},
    \label{loss ce}
\end{equation}
where $y_i$ is the label of node $v_i$ and $p^{(l)}_{iy_i}$ is the predicted probability of node $i$ belonging to the $y_i$-th group (class) in the $l$-th layer.

\subsection{Pseudo-label Guided Open-world Learning}
\revise{To mitigate the lack of novel class labeling information, we generate reliable pseudo-labels to guide the open-world graph learning. The generation process of pseudo-labels is shown in the pseudo-label generator part, while the utilization of pseudo-labels is detailed in the pseudo-label enhanced structure refinement part.}

\subsubsection{Pseudo-label Generator.} \ours{} introduces the pseudo-label generator to mitigate the lack of novel class labeling information. To
fully leverage the rich information captured at different scales, \ours{} ensembles prediction results estimated by multiple stacked prototypical attention networks. However, ensembling multiple group assignment results is non-trivial. Due to the unsupervised nature of novel class discovery, groups from different layers are not aligned, even the number of groups varies from layer to layer.

We overcome these challenges through the proposed padding, aligning and suppressing operation. Given the set of predictions $\{\mathbf{p}^{(1)},...,\mathbf{p}^{(L)} \}$, we first pad them with zero vectors such that they have the same number of prototype groups. Then we align and ensemble predictions from different layers by the Hungarian algorithm \cite{Kuhn1955TheHM}, 
\begin{equation}
    \hat{p}_{ij} = \frac{1}{L}\sum_{k=1}^L{p^{(k)}_{ij_k}},
\end{equation}
where $j_k$ denotes the prototype group ID in the k-th layer that corresponds to the j-th group in the initial layer. $\hat{p}_{ij}$ is the probability of assigning node $i$ to group $j$ obtained from ensemble inference. Groups which only exist in a certain layer or associates with almost no nodes can be regarded as detection noise. We suppress neurons responding to these noisy groups by masking over the prediction results,
\begin{equation}
    \begin{aligned}
        \hat{\mathbf{p}}_i = \hat{\mathbf{p}}_{i} \odot \mathbf{m}\\
        \label{Eq ensemble}
    \end{aligned}    
\end{equation}
where the mask $\mathbf{m}$ is calculated by comparing the group popularity with a pre-defined threshold $\eta$, $m_i = \mathds{1}(\frac{1}{|\mathcal{V}|}\sum_{i}\hat{p}_{i,j}>\eta)$. The symbol $\odot$ stands for the element-wise product. For each known class and discovered novel class, we select reliable pseudo labels as the labeling information on confident node subset, which is the top $\gamma$ percentage of unlabeled nodes with the highest prediction probability (confidence) of belonging to this label. 

\begin{algorithm} 
\LinesNumbered
\caption{\ours{}}
\label{Pseudocode}
\KwData{Graph $G=(\mathcal{V},\mathcal{E},\mathbf{X})$, Labeled training nodes $ (\mathcal{V}_L,\mathcal{Y}^L)$} 
\KwResult{Predictions of unlabeled nodes $\mathcal{V}_U$, where new classes can emerge.}
\BlankLine
Randomly initialize model parameters\;
\Repeat{Convergence or iteration number reaches the threshold}{
    \For{$l=1$ to $L$}{
            Calculate group assignments $\mathbf{p}^{(l)}$ according to Eq. (\ref{Eq groupassign})\;
            Calculate node embeddings $\mathbf{h}^{(l)}$ according to Eq. (\ref{Eq groupatt}) \;}
    Update model parameters according to loss defined in Eq. ($\ref{Eq loss}$)\;
    Update the graph structure according to Eq. (\ref{Eq structure})\;
}
Predict test node labels by ensemble inference in Eq. (\ref{Eq ensemble})\;
\end{algorithm}

\subsubsection{Pseudo-label Enhanced Structure Refinement.} Reliable pseudo-labels information is incorporated into the input graph data for guiding open-world learning. Specifically, we recover intra-class edges to increase similarities between nodes of the same pseudo-label, while eliminating inter-class edges to decrease inter-class similarities. Structure refining according to pseudo-labeling makes it easier to distinguish novel classes from others.

Recovering intra-class edges according to pseudo-labels can introduce information from unconnected nodes in the original graph for open-world learning. However, excessive edge recovery can result in over-smoothing. We select the top $\mu$ percent of intra-class node pairs with the lowest similarities as  recovered edges $\mathcal{E}_{+}$, where the similarity is calculated according to their relationship scores to prototypes in Eq. \ref{Eq n2p},
\begin{equation}
    {s}^n_{ij} = \frac{1}{L}\sum_{l=1}^L \cos(\mathbf{r}^{(l)}_i,\mathbf{r}^{(l)}_j),
\end{equation}
where $\mathbf{r}^{(l)}_i$ is the relationship between nodes and prototypes in the l-th layer. Consequently, edges between intra-class nodes associated to different prototypes are recovered, which increases similarities between prototypes related to the same class and facilitates more accurate semi-supervised prototypical clustering for novel class discovery. 

\ours{} further eliminate inter-class edges $\mathcal{E}_{-}$ according to pseudo-labeling. Since ensemble pseudo-labeling incorporates knowledge from different layers, taking the refined structure as input for each prototypical attention layer can transfer knowledge to eliminate unwanted correlations between classes among layers. The final refined structure can be formulated as, 
\begin{equation}
    \hat{\mathcal{E}}=(\mathcal{E} \setminus \mathcal{E}_{-})\cup\mathcal{E}_{+}.
    \label{Eq structure}
\end{equation}

To make \ours{} robust to the noise introduced during the structure refining process, we introduce the permutation-consistent training loss, which augments the structure information with the adaptive  augmentation technique \cite{Zhu2020GraphCL} and regularizes the prediction consistency between different views,
\begin{equation}
    \mathcal{L}_{con} = \sum_{v_i \in \mathcal{V}}\sum_{l=1}^L \mathrm{KL}(\mathbf{p}^{(l)}_i || {\bar{\mathbf{p}}^{(l)}_i}),
    \label{loss scon}
\end{equation}
where $\bar{\mathbf{p}}_i^{(l)}$ is the prediction over the augmented graph. The overall training loss can be calculated as,
\begin{equation}
\mathcal{L}=\mathcal{L}_{ce}+\mathcal{L}_{reg}+\mathcal{L}_{con}
    \label{Eq loss}
\end{equation}
The final predictions on unlabeled nodes can be obtained based on Eq. (\ref{Eq ensemble}). The training process is summarized in Algorithm \ref{Pseudocode}.

\section{Experiments}
\subsection{Experiment settings}
\renewcommand\arraystretch{1.5}
\begin{table}[t]
    \setlength{\tabcolsep}{1mm}{}
    \centering
    \caption{Statistics of the experiment datasets.}
    \begin{tabular}{c|c|c|c|c|c}
    \hline
    Dataset & \# Nodes & \# Edges & \# Attributes & \# Classes & \# Known Classes\\
    \hline
    Cora & 2708 & 5429 & 1433 & 7 & 5 \\
    \hline
    AmazonPhoto & 7650 & 238162 & 745 & 8 & 6 \\
    \hline
    BlogCatlog & 5196 & 343486 & 8189 & 6 & 4\\
    \hline
    \end{tabular}
    \label{table statistics}
\end{table}

\subsubsection{Datasets.} We test our method on three commonly used datasets:  (1) \textbf{Cora} \cite{Yang2016RevisitingSL} is the citation network where classes are subjects of scientific publications. (2) \textbf{AmazonPhoto} \cite{Shchur2018PitfallsOG} is the co-purchase network with categories of goods as node labels. (3) \textbf{BlogCatlog} \cite{Liu2021AnomalyDO} is the blog network where node labels are topics of blogs. The dataset statistics is shown in Table \ref{table statistics}. For each dataset, we sample 80 percent of classes for which we have labels during training, the remaining ones are novel classes emerged in the test set. We further sub-sample 70 percent and 15 percent of the nodes from known classes to constitute the labelled training set and validation set. The remaining nodes from known classes, along with all instances from novel classes, constitute the unlabeled test node set $V_U$.

\subsubsection{Evaluation Metrics.} Following the novel class discovery literature \cite{liu2023open,vaze2022generalized} in the CV domain, we conduct experiments in scenarios where the number of classes is known as well as in scenarios where it is unknown. The model performance is measured with clustering accuracy, since the label IDs of ground-truth labeling and predictions may not be aligned. Similar to Equation \ref{Eq acc}, the clustering accuracy on test nodes can be calculated as, 
\begin{equation}
\begin{aligned}
\mathrm{ACC} = \mathop{\max}_{f\in \mathcal{P}(\mathcal{Y}^U)}&\sum_{v_i \in \mathcal{V}_U}\frac{1}{|\mathcal{V}_U|}\mathds{1}(\hat{y}_i=f(y_i)). \\
\end{aligned}
\end{equation}
Besides the main metric indicating the classification performance over all testing nodes, we further report values for both known class subset (nodes in $\mathcal{V}_U$
belonging to classes in $\mathcal{Y}^L$) and novel classes subset (nodes in $\mathcal{V}_U$ belonging to classes in $\mathcal{Y}^U \setminus \mathcal{Y}^L$).
\renewcommand\arraystretch{1.5}
\begin{table}[t]
\footnotesize
\caption{Performance comparison of all methods, the experiment results are average over ten runs. ``Prior'' refers to the number of classes.}\label{table main}
\setlength{\tabcolsep}{1.2mm}{}
\begin{tabular}{c|c|c|c|c|c|c|c|c|c}
\hline
\multirow{2}{*}{\textbf{Dataset}}& \multirow{2}{*}{\textbf{Method}} & \multicolumn{4}{c|}{Without Prior} &  \multicolumn{4}{c}{With Prior}\\
\cline{3-10}
& & GCN & GCD & OCD & \ours{} & VGAE & GCD & OCD & \ours{} \\
\cline{2-10}
\hline
\multirow{3}{*}{\textbf{Cora}} & All & 39.29 & 50.67 & 53.28 & \underline{69.91} & 50.60 & 48.08 & 51.03 & \textbf{71.39} \\
\cline{2-10}
& Known & 67.31 & 54.31 & 56.76 & \underline{73.60} & 49.48 & 51.69  & 50.56 & \textbf{76.97} \\
\cline{2-10}
& Novel  &  0.00 & 45.50 & 48.41 & \underline{64.72} & 52.22 & 43.03 & 51.67  & \textbf{65.71}  \\
\cline{2-10}
\hline
\multirow{3}{*}{\textbf{BlogCatalog}} & All & 26.04 & 37.79 & 42.83 & \underline{70.23} &38.18 & 37.91  & 42.73  & \textbf{73.76} \\
\cline{2-10}
& Known & 70.28 & 26.00 & 34.55 & \underline{87.39} & 25.03 & 27.13  & 34.84 & \textbf{88.46}  \\
\cline{2-10}
& Novel & 0.00 & 44.73 & 47.70 & \underline{60.14} & 45.92 & 44.25 & 47.37 & \textbf{65.11} \\
\hline
\multirow{3}{*}{\textbf{\thead{Amazon \\ Photo}}} & All & 56.69 & 70.66 & 67.98 & \underline{84.09} & 66.35 & 70.52  & 67.41  & \textbf{88.05} \\
\cline{2-10}
& Known & 93.44 & 71.98 & 68.58 & \underline{92.71} & 65.89 & 71.98 & 66.75 & \textbf{95.36}  \\
\cline{2-10}
& Novel & 0.00 & 68.44 & 67.04 & \underline{70.79} & 67.06 & 68.27 & 68.43 & \textbf{76.77}  \\
\cline{2-10}

\hline
\multirow{3}{*}{Average} & All & 40.67 & 53.04 & 54.69 & \underline{74.74} & 51.71 & 52.17 & 53.73 & \textbf{77.73}\\
\cline{2-10}
& Known & 77.01 & 50.76 & 53.29 & \underline{84.57} & 46.76 & 50.23 & 50.72 & \textbf{86.93} \\
\cline{2-10}
 & Novel & 0.00 & 52.89 & 54.38 & \underline{65.21} & 54.96 & 51.85 & 55.82 & \textbf{69.20} \\
\hline

\end{tabular}
\end{table}

\subsubsection{Baselines.} We choose the following methods as  baselines, which can be classified into the three families. (1) \textbf{Semi-supervised Approach}. GCN \cite{kipf2016semi} is a popular graph convolution network for semi-supervised node classification, which is unable to discover novel classes and will mistakenly classify novel class nodes into known classes.(2) \textbf{Unsupervised Approach}. VGAE \cite{Kipf2016VariationalGA} is an unsupervised method to learn graph embeddings. We extend this method for open-world graph learning by combining VGAE with the non-parametric kmeans classifier, where the number of classes is assumed to be a known prior. (3) \textbf{Open-world Semi-supervised Approaches.} GCD \cite{vaze2022generalized} is the most popular method for discovering novel categories in the CV domain, which estimates the number of novel classes by maximizing the clustering accuracy on labeled instances. OpenNCD \cite{liu2023open} is the state-of-the-art method for open-world semi-supervised learning. Considering the performance decrease caused by non-parametric classifier in GCD, OpenNCD replaces the kmeans classifier with learnable prototypes and trains the network with a novel progressive bi-level contrastive learning method. We abbreviate OpenNCD as OCD in this article. We extend these methods to open-world graph learning by replacing the encoding module with graph convolution layers \cite{kipf2016semi}. When the number of novel classes is known as a prior, both open-world semi-supervised baselines and \ours{} search for the clustering granularity where the number of clusters (classes) matches with the prior knowledge.

\renewcommand\arraystretch{1.2}
\begin{table}[t]
    \footnotesize
    \setlength{\tabcolsep}{1mm}{}
    \centering
    \caption{Experiment results of class number estimation, where PredNum and MAE represents the estimated class number and mean absolute estimation error.}
    \begin{tabular}{c|c|c|c|c|c|c|c}
    \hline
    \multirow{2}{*}{Method} & \multicolumn{2}{c}{Cora} & \multicolumn{2}{|c|}{Amazon} & \multicolumn{2}{c|}{BlogCatalog} & Average \\
    \cline{2-8}
    & PredNum & MAE & PredNum & MAE & PredNum & MAE & MAE\\
    \hline
    GCD & 6.0 & 1.0 & 7.1 & 0.9 & 6.3 & \underline{0.3} & 0.7\\
    \hline
    OpenNCD & 7.4 & \textbf{0.4} & 8.8 & \underline{0.8} & 5.6 & 0.4 & \underline{0.5}\\
    \hline
    \ours{} & 7.7 & \underline{0.7} & 8.0 & \textbf{0.0} & 5.7 & \textbf{0.3} & \textbf{0.3}\\
    \hline
    \end{tabular}
    \label{table number}
\end{table}


\renewcommand\arraystretch{1.2}
\begin{table}[t]
    \setlength{\tabcolsep}{1.5mm}{}
    \centering
    \footnotesize
    \caption{Ablation study, where PAN stands for prototypical attention network and POL stands for pseudo-label guided open-world learning.}
    \begin{tabular}{cc|c|c|c|c|c|c|c|c}
    \hline
    \multicolumn{2}{c|}{Modules}& \multicolumn{4}{c}{Without prior} & \multicolumn{4}{|c}{With prior}\\
    \cline{1-10}
      \multirow{2}{*}{PAN} &\multirow{2}{*}{POL} & \multicolumn{2}{c|}{Cora} & \multicolumn{2}{c|}{BlogCatalog} & \multicolumn{2}{c|}{Cora} & \multicolumn{2}{c}{BlogCatalog} \\
     \cline{3-10}
      &  &  All & Novel &  All & Novel &  All & Novel &  All & Novel \\
    \hline
    $\times$ & $\times$ & 53.28 & 48.41 & 42.83 & 47.70 & 51.03 & \underline{51.67} & 42.73 & 47.27 \\
    \hline
    \checkmark & $\times$ & \underline{65.97} & \underline{55.27} & \textbf{70.44} & \textbf{60.38} & \underline{64.53} & 49.10 & \underline{73.10} &  \underline{63.80}\\
    \hline
    \checkmark & \checkmark & \textbf{69.91}  & \textbf{64.72} & \underline{70.23} & \underline{60.14} & \textbf{71.39} &  \textbf{65.71} & \textbf{73.76} & \textbf{65.11} \\
    \hline
    
    \end{tabular}
    \label{table ablation}
\end{table}
    
\subsection{Main results}
\begin{figure}[t]
    \centering
    \includegraphics[width=\linewidth]{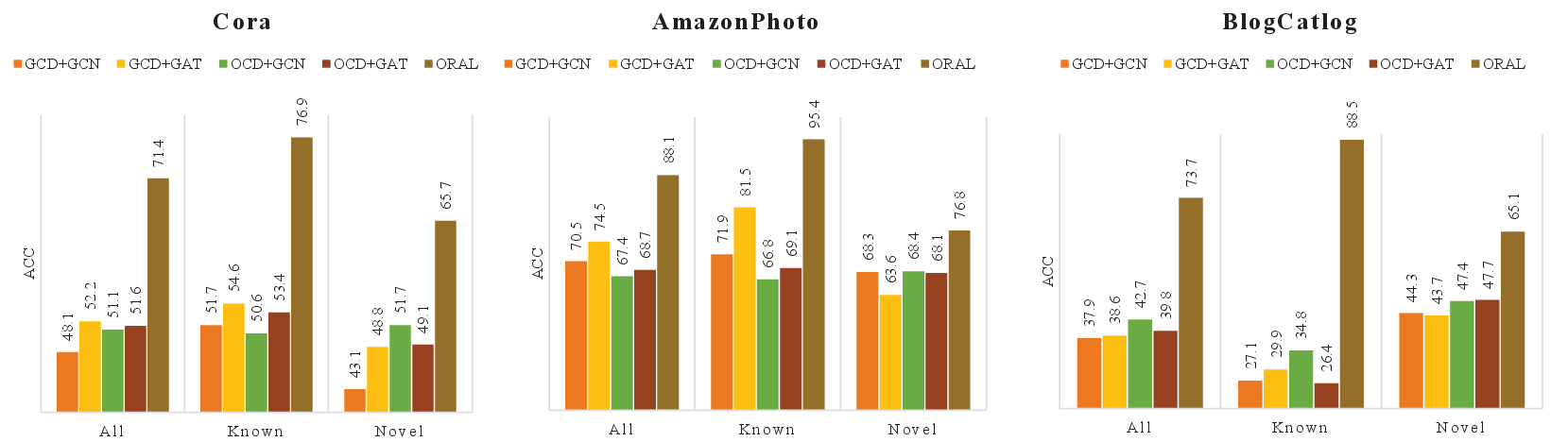}
    \caption{Performance of baselines with different graph convolution networks. Our \ours{} consistently outperforms all variants of existing open-world baseline methods, showing its superiority.}
    \label{Figure encoder}
\end{figure}

Table \ref{table main} illustrates the experiment results of baselines and our work. From the results, we have the following observations: 
\textbf{(1)} Our proposed method consistently shows strong performance on both known and novel classes across all datasets. The improvements  can be attributed to that \ours{} distinguishes novel classes from others by utilizing node similarities discovered by semi-supervised
prototypical clustering and generates reliable pseudo-labels to mitigate lack of novel class labeling information. 
\textbf{(2)} Despite existing semi-supervised method GCN performs well on known classes, they are unable to discover novel classes and mistakenly classify novel class nodes into known classes. Consequently, its accuracy over novel classes is zero.
\textbf{(3)} Given the number of classes, we can approximate the ground-truth labels by combining clustering and unsupervised representation learning method VGAE, which results in low performance on known classes due to the lack of supervision information.
\textbf{(4)} With the help of labeled known class nodes, open-world semi-supervised learning methods can discover novel classes automatically without knowing the number of classes in advance. \ours{} achieves the lowest mean absolute error in the estimated class number as shown in Table \ref{table number}. 
\textbf{(5)} The performance of open-world semi-supervised learning baselines is still unsatisfactory since they are unable to eliminate correlations between classes on graphs. We further compare the performance of baselines equipped with different graph convolution methods in Figure \ref{Figure encoder}. Although graph attention network \cite{Velickovic2017GraphAN} can assign learnable scores to edges, it still lacks the knowledge to distinguish novel classes from known ones.
\textbf{(6)} By guiding the models to search for the proper labeling granularity during training, the number of novel classes can improve the novel class accuracy of OpenNCD and \ours{} by 1.5\% and 4.0\% separately. Nonetheless, the prior information is simply incorporated into the clustering during testing in GCD, which can not guide the learning process and improve performance \cite{Wen2022ParametricCF}.
\textbf{(7)} Due to the lack of novel class labeling information, open-world semi-supervised  baselines generally perform better on known classes than novel ones, except for the BlogCatlog dataset. Baseline methods tend to cluster different classes into the same group on BlogCatlog due to the strong inter-class correlations. To achieve best overall performance, the clustering accuracy is biased towards classes with more testing nodes, which constitute the novel classes in BlogCatlog.

\begin{figure}[t]
    \centering
    \includegraphics[width=0.95\linewidth]{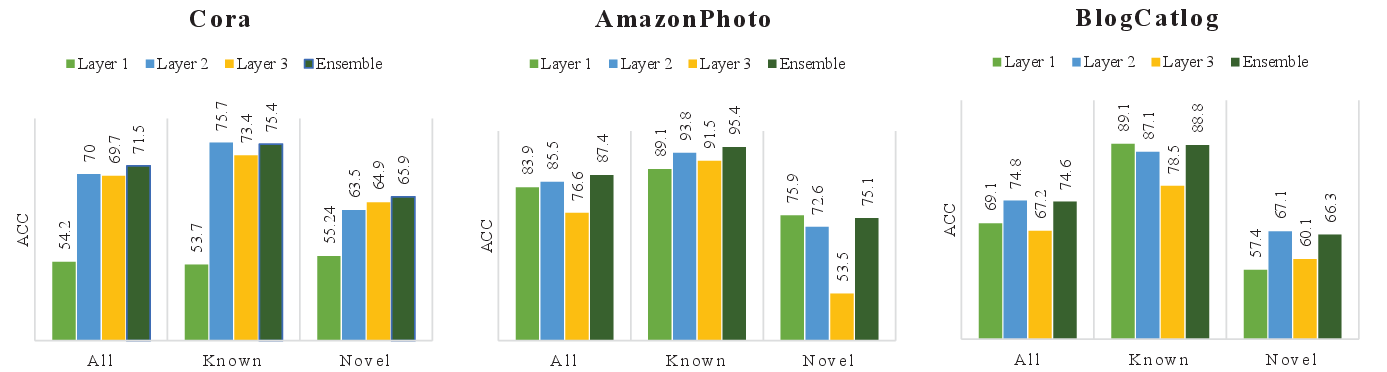}
    \caption{Performance of different prototypical attention network layers.}
    \label{Figure layer}
\end{figure}

\begin{figure}[t]
    \centering
    \includegraphics[width=0.95\linewidth]{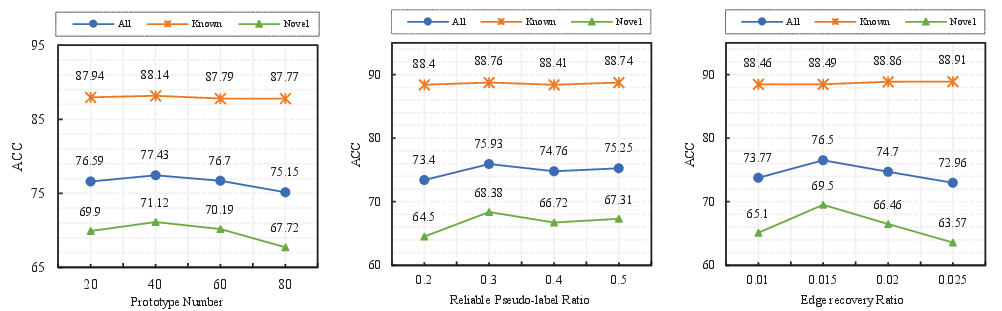}
    \caption{Impact of key hyper-parameters on performance of \ours{} on the BlogCatlog dataset when the class number is known.}
    \label{Figure sens}
\end{figure}

\subsection{Abaltion Study}
In this section, we investigate the contribution of two main components in our model by ablation experiments. Specifically, we ablate the prototypical attention network (PAN) by replacing it with the graph convolution layer \cite{kipf2016semi}. The ablation of the pseudo-label guided open-world learning (POL) module is achieved by directly taking the model trained on the original graph for prediction, where the incorporation of pseudo-labeling information is ablated. The results of these variants are shown in Table \ref{table ablation}, we have the following observations: \textbf{(1)} The improvement of PAN module is more significant on BlogCatlog which has more inter-class edges (60\%) compared with Cora (19\%). This observation shows the effectiveness of PAN in eliminating inter-class correlations on graphs. \textbf{(2)} Adding the POL module can further improve accuracy over novel classes, indicating the effectiveness of POL in mitigating the lack of novel class labeling information.

We also conduct experiments to show the effectiveness of the proposed ensemble inference technique in Eq. \ref{Eq ensemble}. The performance of different prototypical attention network layers is shown in Figure \ref{Figure layer}, we have the following observations: \textbf{(1)} The layer which achieves the best performance varies across datasets and classes. For instance, the second layer exhibits optimal performance for known classes classification in the AmazonPhoto dataset, whereas it is surpassed by the third layer on novel classes in the Cora dataset. \textbf{(2)} The ensemble inference consistently achieves strong performance over different classes, even surpassing any individual layer. Moreover, it alleviates the challenge of manually searching for the optimal layer for prediction, particularly in cases where labeled validation data for novel classes is unavailable.



\subsection{Impact of Hyper-parameter Settings} 
Next, we conduct experiments to investigate the impact of key hyper-parameters on the performance of \ours{}:

\noindent \textbf{Impact of prototype number $N_{pro}$}. Prototypes are representative nodes on graphs, which are modeled as trainable vectors. We select \{20, 40, 60, 80\} as $N_{pro}$ respectively, and show experiment results in Figure \ref{Figure sens}. The performance improves as $N_{pro}$ increases and peaks at 40. Then the performance decreases as $N_{pro}$ becomes larger. Insufficient number of prototypes are not representative of nodes on graphs,  while setting $N_{pro}$ too high can lead to over-fitting, both results in performance drops.

\noindent \textbf{Impact of reliable pseudo-label ratio $\gamma$}. The pseudo-label generator selects the labeling information on top $\gamma$ percentage of unlabeled nodes with the highest prediction probability (confidence) as reliable pseudo-labels. We range $\gamma$ within \{0.2, 0.3, 0.4, 0.5\} and plot the results. In the beginning, performance increases as $\gamma$ increases, indicating the effectiveness of utilizing pseudo-labels to guide open-world learning. The performance declines when $\gamma$ is larger than 0.3, which may be caused by noise in pseudo-labels with low prediction confidence. 

\noindent \textbf{Impact of recovery edge ratio $\mu$}. \ours{} selects top $\mu$ percent of intra-class node pairs with the lowest similarities as recovered edges to guide open-world learning. We select $\mu$ as \{0.01, 0.015, 0.02, 0.025\} respectively. When $\mu$ is 0.015, \ours{} achieves the best performance. This is because too many edges can lead to over-smoothing, while recovering too few edges can not incorporate enough information into the input graph to guide open-world learning.


\section{Related work}
In this section, we briefly introduce the relevant research lines of our work, namely graph neural networks and novel class discovery.

\noindent \textbf{Graph Neural Networks.} Graph Neural Networks (GNNs) are designed to use deep learning architectures on graph-structured data \cite{Gori2005ANM,Scarselli2009TheGN,Wu2020UnsupervisedDA} and node classification \cite{Bruna2013SpectralNA,huang2018adaptive,Izadi2020OptimizationOG,kipf2016semi,luan2021heterophily,biology,Niepert2016LearningCN} is one of the most important downstream tasks of GNNs. Early methods mainly focus on classifying the nodes within a set of fixed classes with abundant labelling \cite{Bruna2013SpectralNA,Niepert2016LearningCN}. To ease the annotation burden, researchers have paid much attention to semi-supervised setting \cite{huang2018adaptive,kipf2016semi} and few-shot setting \cite{Ding2020GraphPN,Lu2022GeometerGF}. However, these methods are unable to automatically discover novel classes emerged in the unlabeled test data where the method is required to estimate the number of novel classes and classify the unlabeled nodes into either known or novel classes.

\noindent \textbf{Novel Class Discovery.} To handle the unknown novel classes during testing, early works \cite{Geng2018RecentAI,Sun2020ConditionalGD,Wu2020OpenWGLOG,Zhang2022ADV} propose out-of-distribution detection methods to reject samples not belonging to any known classes based on the statistics of prediction logits. However, these methods can not further discover the novel classes among the rejected samples. Under the assumption that the number of novel classes is known, \cite{Brbic2020MARSDN,Han2019LearningTD,Hsu2017LearningTC,Zhong2020OpenMixRK} propose to discover novel classes by clustering unlabeled data on the representation space with the help of labeled known classes. \cite{liu2023open,vaze2022generalized,Zhao2023LearningSG} further propose to automatically estimate the novel class number by utilizing the labeling granularity provided by known classes. Nonetheless, all these methods face challenges in distinguishing novel classes from others on graphs due to the unwanted correlations between classes caused by edge connections. 
\section{Conclusion}
\revise{In this paper, we propose \ours{} to discover novel classes for open-world graph learning. As far as we known, this is the first work to deal with this challenging yet practical problem. \ours{} first employs the prototypical attention network to distinguish novel classes from others during representation learning. \ours{} further generates pseudo-labels through clustering ensembles to exploit unlabeled data and mitigate the lack of novel class labeling information. Extensive experiment results on three public datasets show the superiority of our proposed method. In the future, we would like to extend our method to more challenging scenarios, like discovering novel classes on class-imbalanced graphs.}
\bibliographystyle{splncs04}
\bibliography{main}

\begin{thebibliography}{10}
\providecommand{\url}[1]{\texttt{#1}}
\providecommand{\urlprefix}{URL }
\providecommand{\doi}[1]{https://doi.org/#1}

\bibitem{Brbic2020MARSDN}
Brbi{\'c}, M., Zitnik, M., Wang, S., et~al.: Mars: discovering novel cell types across heterogeneous single-cell experiments. Nature Methods  \textbf{17},  1200--1206 (2020)

\bibitem{Bruna2013SpectralNA}
Bruna, J., Zaremba, W., Szlam, A., et~al.: Spectral networks and locally connected networks on graphs. arXiv preprint arXiv:\href{https://arxiv.org/abs/1312.6203}{1312.6203}  (2013)

\bibitem{Cao2021OpenWorldSL}
Cao, K., Brbic, M., Leskovec, J.: Open-world semi-supervised learning. In: Proceedings of ICLR (2022)

\bibitem{Chen2020ConvolutionalKN}
Chen, D., Jacob, L., Mairal, J.: Convolutional kernel networks for graph-structured data. In: Proceedings of ICML (2020)

\bibitem{Ding2020GraphPN}
Ding, K., Wang, J., Li, J., et~al.: Graph prototypical networks for few-shot learning on attributed networks. In: Proceedings CIKM (2020)

\bibitem{Gao2019ExploringSG}
Gao, X., Hu, W., Guo, Z.: Exploring structure-adaptive graph learning for robust semi-supervised classification. In: Proceedings of ICME (2020)

\bibitem{Geng2018RecentAI}
Geng, C., Huang, S.J., Chen, S.: Recent advances in open set recognition: A survey. In: TPAMI (2020)

\bibitem{Gori2005ANM}
Gori, M., Monfardini, G., Scarselli, F.: A new model for learning in graph domains. In: Proceedings of IJCNN. IEEE (2005)

\bibitem{Han2019LearningTD}
Han, K., Vedaldi, A., Zisserman, A.: Learning to discover novel visual categories via deep transfer clustering. In: Proceedings of ICCV (2019)

\bibitem{Hetzel2021GraphRL}
Hetzel, L., Fischer, D.S., G{\"u}nnemann, S., et~al.: Graph representation learning for single-cell biology. Current Opinion in Systems Biology  (2021)

\bibitem{Hsu2017LearningTC}
Hsu, Y.C., Lv, Z., Kira, Z.: Learning to cluster in order to transfer across domains and tasks. In: Proceedings of ICLR (2018)

\bibitem{huang2018adaptive}
Huang, W., Zhang, T., Rong, Y., et~al.: Adaptive sampling towards fast graph representation learning. In: Proceedings of NeurlIPS  (2018)

\bibitem{Izadi2020OptimizationOG}
Izadi, M.R., Fang, Y., Stevenson, R.L., et~al.: Optimization of graph neural networks with natural gradient descent. In: Proceedings of IEEE Big Data (2020)

\bibitem{Kipf2016VariationalGA}
Kipf, T.N., Welling, M.: Variational graph auto-encoders. In: Proceedings of NeurIPS (2016)

\bibitem{kipf2016semi}
Kipf, T.N., Welling, M.: Semi-supervised classification with graph convolutional networks. In: Proceedings of ICLR (2017)

\bibitem{Kuhn1955TheHM}
Kuhn, H.W.: The hungarian method for the assignment problem. In: NRL  (1955)

\bibitem{Liao2018GraphPN}
Liao, R., Brockschmidt, M., Tarlow, D., et~al.: Graph partition neural networks for semi-supervised classification. arXiv preprint arXiv:\href{https://arxiv.org/abs/1803.06272}{1803.06272}  (2018)

\bibitem{liu2023open}
Liu, J., Wang, Y., Zhang, T., et~al.: Open-world semi-supervised novel class discovery. In: Proceedings of IJCAI (2023)

\bibitem{Liu2021AnomalyDO}
Liu, Y., Li, Z., Pan, S., et~al.: Anomaly detection on attributed networks via contrastive self-supervised learning. In: IEEE TNNLS  (2021)

\bibitem{Lu2022GeometerGF}
Lu, B., Gan, X., Yang, L., et~al.: Geometer: Graph few-shot class-incremental learning via prototype representation. In: Proceedings of SIGKDD (2022)

\bibitem{luan2021heterophily}
Luan, S., Hua, C., Lu, Q., et~al.: Is heterophily a real nightmare for graph neural networks on performing node classification? arXiv preprint arXiv:\href{https://arxiv.org/abs/2109.05641}{2109.05641}  (2021)

\bibitem{biology}
Muzio, G., O’Bray, L., Borgwardt, K.: Biological network analysis with deep learning. In: Briefings in bioinformatics  (2021)

\bibitem{Niepert2016LearningCN}
Niepert, M., Ahmed, M., Kutzkov, K.: Learning convolutional neural networks for graphs. In: Proceedings of ICML (2016)

\bibitem{Scarselli2009TheGN}
Scarselli, F., Gori, M., Tsoi, A.C., et~al.: The graph neural network model. In: IEEE Trans. Neural Netw.  (2008)

\bibitem{Shchur2018PitfallsOG}
Shchur, O., Mumme, M., Bojchevski, A., et~al.: Pitfalls of graph neural network evaluation. arXiv preprint arXiv:\href{https://arxiv.org/abs/1811.05868}{1811.05868}  (2018)

\bibitem{Sun2020ConditionalGD}
Sun, X., Yang, Z., Zhang, C., et~al.: Conditional gaussian distribution learning for open set recognition. In: Proceedings of CVPR (2020)

\bibitem{vaze2022generalized}
Vaze, S., Han, K., Vedaldi, A., et~al.: Generalized category discovery. In: Proceedings of CVPR (2022)

\bibitem{Velickovic2017GraphAN}
Velickovic, P., Cucurull, G., Casanova, A., et~al.: Graph attention networks. In: Proceedings of ICLR (2018)

\bibitem{Wen2022ParametricCF}
Wen, X., Zhao, B., Qi, X.: Parametric classification for generalized category discovery: A baseline study. In: Proceedings of ICCV (2023)

\bibitem{Wu2020UnsupervisedDA}
Wu, M., Pan, S., Zhou, C., et~al.: Unsupervised domain adaptive graph convolutional networks. In: Proceedings of WWW (2020)

\bibitem{Wu2020OpenWGLOG}
Wu, M., Pan, S., Zhu, X.: Openwgl: Open-world graph learning. In: Proceedings of ICDM (2020)

\bibitem{Yang2016RevisitingSL}
Yang, Z., Cohen, W., Salakhudinov, R.: Revisiting semi-supervised learning with graph embeddings. In: Proceedings of ICML (2016)

\bibitem{Zhang2022ADV}
Zhang, Q., Li, Q., Chen, X., et~al.: A dynamic variational framework for open-world node classification in structured sequences. In: Proceedings of ICDM (2022)

\bibitem{Zhao2023LearningSG}
Zhao, B., Wen, X., Han, K.: Learning semi-supervised gaussian mixture models for generalized category discovery. In: Proceedings of ICCV (2023)

\bibitem{Zhong2020OpenMixRK}
Zhong, Z., Zhu, L., Luo, Z., et~al.: Openmix: Reviving known knowledge for discovering novel visual categories in an open world. In: Proceedings of CVPR (2021)

\bibitem{Zhu2020GraphCL}
Zhu, Y., Xu, Y., Yu, F., et~al.: Graph contrastive learning with adaptive augmentation. In: Proceedings of WWW (2021)

\end{thebibliography}
%




\end{document}